\documentclass[pdflatex,sn-mathphys-num]{sn-jnl}

\usepackage{graphicx}%
\usepackage{multirow}%
\usepackage{amsmath,amssymb,amsfonts}%
\usepackage{amsthm}%
\usepackage{mathrsfs}%
\usepackage[title]{appendix}%
\usepackage{xcolor}%
\usepackage{textcomp}%
\usepackage{manyfoot}%
\usepackage{booktabs}%
\usepackage{algorithm}%
\usepackage{algorithmicx}%
\usepackage{algpseudocode}%
\usepackage{listings}%

\begin{document}

\title[Article Title]{Tran-GCN: A Transformer-Enhanced Graph
Convolutional Network for Person
Re-Identification in Monitoring Videos}


\author[1,3]{\fnm{Xiaobin} \sur{Hong}}\email{xiaobin20@graduate.utm.my}

\author[1]{\fnm{Tarmizi} \sur{Adam}}\email{Tarmizi.adam@utm.my}

\author[2]{\fnm{Masitah} \sur{Ghazali}}\email{Masitah@utm.my}

\affil[1]{\orgdiv{Faculty of Computing}, \orgname{Universiti Teknologi Malaysia}, \orgaddress{\street{Jalan Iman}, \city{Skudai}, \postcode{81310}, \state{Johor}, \country{Malaysia}}}

\affil[2]{\orgdiv{Malaysia-Japan International Institute of Technology}, \orgname{Universiti Teknologi Malaysia}, \orgaddress{\street{Kuala Lumpur}, \postcode{54100},  \country{Malaysia}}}


\abstract{Person Re-Identification (Re-ID) has gained popularity in computer vision, enabling cross-camera pedestrian recognition. Although the development of deep learning has provided a robust technical foundation for person Re-ID research, most existing person Re-ID methods overlook the potential relationships among local person features, failing to adequately address the impact of pedestrian pose variations and local body parts occlusion. 
Therefore, we propose a Transformer-enhanced Graph Convolutional Network (Tran-GCN) model to improve Person Re-Identification performance in monitoring videos. The model comprises four key components: (1) A Pose Estimation Learning branch is utilized to estimate pedestrian pose information and inherent skeletal structure data, extracting pedestrian key point information; (2) A Transformer learning branch learns the global dependencies between fine-grained and semantically meaningful local person features; (3) A Convolution learning branch uses the basic ResNet architecture to extract the person's fine-grained local features; (4) A Graph Convolutional Module (GCM) integrates local feature information, global feature information, and body information for more effective person identification after fusion.
Quantitative and qualitative analysis experiments conducted on three different datasets (Market-1501, DukeMTMC-ReID, and MSMT17) demonstrate that the Tran-GCN model can more accurately capture discriminative person features in monitoring videos, significantly improving identification accuracy.}

\keywords{Person Re-Identification,ResNet, OpenPose, Graph Convolutional Network}



\maketitle

\section{Introduction}\label{sec1}

Person Re-Identification is a technology that utilizes computer vision techniques to determine whether a specific pedestrian exists in an image or video sequence, aiming to retrieve individuals of interest through multiple non-overlapping cameras \cite{bib1,sezavar2024new}. As a fundamental computer vision task, Person Re-Identification is used in various fields such as public safety, traffic management, smart urban management and autonomous driving \cite{gong2024tclanenet, zhang2023skipcrossnets, song2023graphalign++}.

As Person Re-Identification evolves rapidly, deep learning has emerged as the primary research and technical approach in this domain\cite{yadav2024deep}.Yi and Li et al. \cite{bib2, bib3} introduce Person Re-Identification methods grounded in feature representation, utilizing Siamese network models to compute similarities between pedestrian images. Weinberger et al. \cite{bib4} propose a metric learning-based approach, designing a suitable and effective triplet loss function through a function to ensure that pedestrian images adhere to specific distributions in the target feature space. Wang, Song, and Zhao et al. \cite{bib5, bib6, bib7} introduce methods centered on local features, adopting grid partitioning, human semantic segmentation techniques, and utilizing pedestrian key points to extract local features from specific body parts. Unfortunately, most existing Person Re-Identification methods overlook the inherent structural information of the human body and the relationships between local features. This can result in inaccurate retrieval outcomes when pedestrian poses vary or when pedestrians exhibit similar appearances. The inadequate handling of pedestrian poses and the relationships between local features has limited the generalization capabilities of models, thereby hindering the progress of Person Re-Identification technology.

At present, several deep learning models excel in extracting local pedestrian features and estimating pedestrian poses. For example, part-based methods \cite{bib8, hu2021railway, bib9, bib10} adopt horizontal or grid partitioning to obtain various local regions of pedestrians, Using ResNet to obtain global features,facilitating the alignment of pedestrian images \cite{song2024contrastalign}. However, due to the small size of pedestrian datasets, these models are prone to overfitting. Metric learning-based methods \cite{bib11, bib12, li2023tvg} emphasize designing suitable and effective metric loss functions to continuously reduce the distance between images of the same pedestrian and increase the distance between images of different pedestrians. Although this approach achieves high recognition accuracy, its generalization performance suffers when confronted with complex backgrounds. Local feature-based methods \cite{bib13, bib14, bib15, gong2023feature} utilize manual partitioning or auxiliary information such as pedestrian poses to acquire local regional features from pedestrian images, thereby mitigating the effects of misalignment and pose variations. Nevertheless, these methods overlook the potential relationships between local pedestrian features, failing to extract discriminative key information when noise is present in images.
Multimodal fusion optimization is verified by demonstrating improved performance metrics, such as accuracy and robustness, when integrating data from multiple modalities compared to using individual modalities alone \cite{zhang2023multi, gong2023sifdrivenet, song2023graphalign}.
Therefore, it is crucial for person re-identification networks to comprehensively consider the extraction of local features, the global dependencies between these local features, and pedestrian poseture or gesture information \cite{ertugrul2020attaining, zeghoud2022real}.

In this paper, we propose a Tran-GCN person re-identification model that systematically integrates local features, the global dependencies between these features, and pedestrian pose information using a graph convolution module, thereby capturing more detailed and comprehensive pedestrian characteristics to significantly enhance the performance and accuracy of person re-identification.
First, pedestrian pose estimation is utilized to obtain rich joint skeleton structure information of pedestrians, constructs a topological graph of pedestrian joints, extracts local features of each key point using joint information, and learns the adjacency matrix of the human topology graph. Next, ResNet and Transformer extract fine-grained local features and global appearance features relationship of pedestrians, respectively. Finally, GCM integrates the correlations among these three types of features to extract more discriminative pedestrian features, effectively addressing challenges such as variations in pedestrian poses, similarities in pedestrian appearances, and partial occlusions.

The main contributions of this study are as follows:
\begin{itemize}

\item  
We present a Tran-GCN network, which simultaneously captures local features, global dependencies of local features, and pedestrian pose information, and then uses graph networks to generate more discriminative features, enhancing the ability to distinguish between different pedestrians.

\item  A graph convolution module is proposed to effectively and efficiently integrate local features, global features, and pedestrian pose information by capturing features through the adjacency matrix, enabling a high-level understanding of pedestrian characteristics and behaviors.

\item  Extensive experiments on the Market-1501 \cite{bib16}, DukeMTMC-ReID \cite{bib17}, and MSMT17 \cite{bib18} Person Re-Identification datasets demonstrate the effectiveness and robustness of the proposed Tran-GCN model in addressing various challenges in Person Re-Identification, highlighting its outstanding performance across these diverse datasets.
\end{itemize}

\section{Related work}\label{sec2}
In recent years, deep learning-based methods have dominated the field of Person Re-Identification. In this section, we comprehensively review three popular Person Re-Identification approaches: feature representation-based methods, metric learning-based methods, and local feature-based methods. These approaches are closely related to our proposed Person Re-Identification method.

\subsection{Feature Representation-based Methods}

Most existing Person Re-Identification techniques focus on this approach, leveraging the concept of image classification tasks to transform Person Re-Identification into either a classification or verification task. Typically, the entire pedestrian image was input into a network to extract the global features of the pedestrian. Geng et al. \cite{bib19} aimed to fully utilize pedestrian label information by utilizing a joint learning approach with a classification subnetwork and a verification subnetwork, thereby extracting more discriminative pedestrian features. In addition to utilizing pedestrian label information, various attribute label information of pedestrians, such as long or short hair, whether carrying a backpack, whether wearing a hat, etc., had also been fully explored. For example, Zhang et al.  \cite{bib20} proposed a multi-task learning framework that combined attribute label information and jointly trained classification and verification networks to extract complementary and discriminative pedestrian features. Ahmed et al. \cite{bib21} added a cross-input neighborhood difference layer and an image patch difference accumulation layer to the Siamese network. By calculating the differences between adjacent local region blocks, they were able to obtain the similarity between pedestrian images. Although this method could extract complementary and discriminative pedestrian features, the small size of pedestrian datasets could easily lead to model overfitting.

\subsection{Metric Learning-based Methods}

This Person Re-Identification method focused on designing appropriate and effective metric loss functions to reduce the intra-class distance (between similar pedestrians) and increase the inter-class distance (between different pedestrians). Cheng et al. \cite{bib22} proposed the Triplet Loss function, which involved inputting three pedestrian images at a time: an anchor image, a positive sample (similar to the anchor), and a negative sample (dissimilar to the anchor). During the network optimization process, the distance between positive sample pairs is minimized, while the distance between negative sample pairs is maintained above a certain threshold.To enhance the performance of the Triplet Loss, Hermans et al. \cite{bib23} introduced the concept of Hard Sample Mining into the Triplet Loss function. This improvement involved selecting, within each input batch, the least similar positive sample and the most similar negative sample to the anchor sample as "hard samples" during training. This approach helped improve the model's generalization ability.Furthermore, Chen et al. \cite{bib24} proposed the Quadruplet Loss function, which extended the Triplet Loss by adding a pair of negative samples. This additional pair served as a weakly supervised term, enhancing the model's learning capability. While this method achieved high recognition accuracy, the model's generalization performance may decrease when there were significant variations in pedestrian poses.

\subsection{Local Features-based Methods}

Recently, numerous studies have leveraged local feature information to improve the representation of local features in Person Re-Identification tasks. Some part-based methods  \cite{bib25,bib26,bib27} utilized manually defined horizontal or network-partitioned pedestrian local regions, which tended to be coarse and did not account for the refined local characteristics of pedestrians. These methods were sensitive to changes in pedestrian poses. On the other hand, mask-based methods \cite{bib28,bib29,bib30,bib31,tan2023dmdsnet} could effectively eliminate the impact of background noise, but they primarily contain shape information of the human body, and the extracted semantic information heavily relied on the performance of semantic segmentation algorithms. Pose-based methods 
 \cite{bib32,bib33,tan2023lightweight,bib34}, while capable of mitigating the effects of pedestrian misalignment and pose variations, often overlooked the relationships between local features of pedestrians. When there were partial occlusions in the images, these methods might lose some of their generalization capabilities.

\section{Proposed method}
\label{sec:proposed_method}

In this section, we first introduce the overall structure of the network, followed by a description of the multi-branch feature extraction backbone. Next, we explain how the GCM efficiently and effectively integrates the multiple features extracted by the multi-branch Backbone. Finally, the loss function for training the multi-branch network is discussed in Section 3.4.

\subsection{Overview}
\label{Overview}

\begin{figure}[!t]
  \centering
  \includegraphics[width=12.5cm, height=9.5cm]{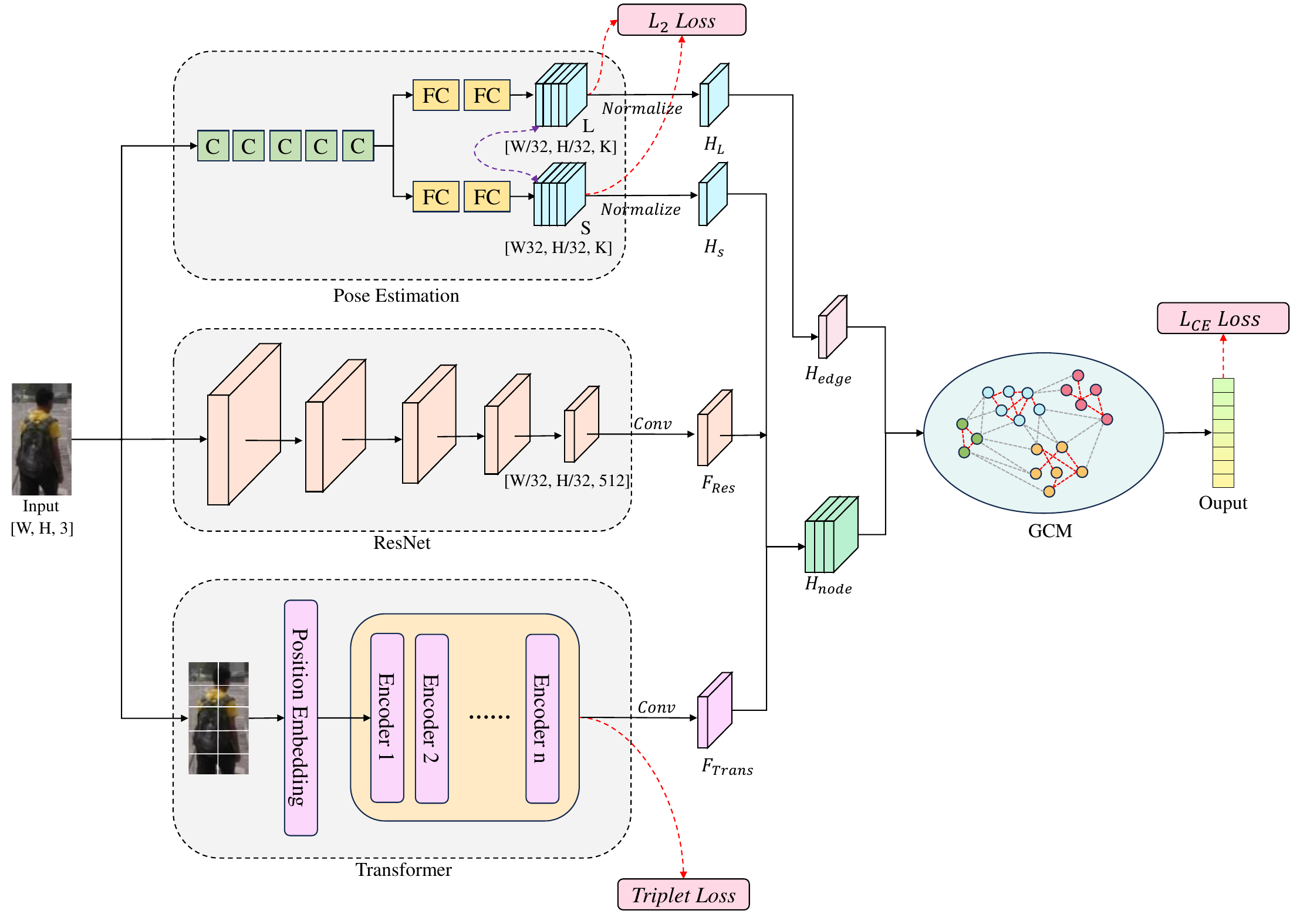}
 \caption{Illustration of our proposed framework includes two parts: (1) Multi-branch Feature Extraction Backbone which extracts pedestrian multi-scale features; (2) GCM branch which performs fusing the pedestrian features from above.}
  \label{fig:fig1}
\end{figure}

Fig. \ref{fig:fig1} illustrates the overall network structure of the proposed Tran-GCN, which comprises two main parts: the multi-branch Feature Extraction backbone and the feature fusion Graph Convolutional module (GCM). We denote the input as a probe person image. This probe image Input is passed through the Multi-Branch Feature Extraction backbone to obtain the pedestrian's pose information, global features, and local feature relationships. The first part contains three components, each with a corresponding loss function, explained in Section 3.2 and Section 3.4. Finally, GCM integrates the above pedestrian information to generate richer and more discriminative feature representations, aiding in accurate pedestrian identification in complex backgrounds, which is explained in detail in Section 3.3.

\subsection{Multi-Branch Feature Extraction Backbone}
The section includes three components: (1) Pose Estimation Module: this component provides data on the positions of the pedestrian's skeleton and joints. (2) ResNet Module: this component extracts fine-grained local features, such as clothing textures, colors, and other subtle markers; (3) Transformer Module: this component establishes global dependencies between different local features, making it less susceptible to local occlusions and partial deformations.

\subsubsection{Pose Estimation Learning Module}

Evaluating whole-body motion is challenging because of the articulated nature of the skeleton structure \cite{bib35}. Pedestrian pose information provides data on the positions of the skeleton and joints, offering fine-grained features for pedestrians with varying poses and partial occlusions, thereby enhancing re-identification performance to handle various complex scenarios and challenges.Therefore, we adopt a pedestrian pose detection network to obtain the positions of the keypoints and the connections between the skeletons. 

Specifically, the input image $I \in \mathbb{R}^{W\times H \times 3}$ is fed into VGG-16 to extract deep feature representations $F_{pe} \in \mathbb{R}^{ \frac{W}{32} \times \frac{H}{32} \times 512}$, followed by simultaneous prediction of confidence maps and affinity fields,,it can be mathematically denoted by the following equation. In which, the $W$, $H$, and $3$ represent the width, height, and channels of the input image.
\begin{equation}
    F_{pe} =VGG16(I).
\end{equation}
In the pose estimation section of Fig. \ref{fig:fig1}, $L \in \mathbb{R}^{\frac{W}{32} \times \frac{H}{32} \times K}$ and $S \in \mathbb{R}^{\frac{W}{32} \times \frac{H}{32} \times K}$ represent the affinity fields and confidence maps, respectively. The $K$ represents the key point in the feature map.
Since confidence maps provide information about the likelihood of keypoints at each location, helping to identify the most probable positions of keypoints, and affinity fields capture the relationships between different keypoints, indicating the degree of association between pairs of keypoints (e.g., those connecting bones) and aiding in determining the connections between keypoints to construct the skeleton structure. Therefore, we incorporate information from affinity fields when predicting confidence maps, and vice versa.

\begin{equation}
\begin{aligned}
& S=\rho \left(F_{pe}, L\right), \\
& L=\phi \left(F_{pe}, S\right),
\end{aligned}
\end{equation}

\noindent
where $\rho$(,) and $\phi$(,) are the fully connections layer having import $F_{pe}$,$L$ and $S$ respectively. Compared with the original OpenPose \cite{bib35}, the backbone of generating $L$ and $S$ shares weights to reduce the parameter count by half. Additionally, we have removed the multi-stage iterative steps from OpenPose to improve computational efficiency and speed. For complex pose variations and occlusion scenarios, these can be addressed subsequently through the GCM. 

To obtain rich pose information of pedestrians, we pre-trained OpenPose on the COCO dataset \cite{bib36}, which extracts $K$=18 keypoint heatmaps and $N$=19 part affinity fields from the input pedestrian images. Each generated keypoint heatmap $S$ and confidence score $L$ contain information about the pedestrian's joints. 

\subsubsection{ResNet Learning branch}

Due to the fact that each convolutional kernel (filter) in a Convolution Neural Netowrk (CNN) operates only within a local region and progressively moves across the entire image through a sliding window approach, CNN possesses the ability to capture local features such as edges, corners, and textures. In the task of Person Re-Identification, local features are crucial for identifying and distinguishing different individuals. Especially in complex backgrounds, the focus on local features enhances the model's robustness. Therefore, we employ a ResNet-50 pre-trained on ImageNet \cite{bib37} as the backbone network.

To adapt the network for pedestrian feature extraction, the final average pooling layer and fully connected layer of the original ResNet-50 are removed. The stride of the convolutional layer in the first residual block of Stage 4 in the ResNet-50 structure is set to 1, modifying the network architecture to better suit the specific task of pedestrian feature extraction.
\begin{equation}
    F_{Res} = ResNet50(I),
\end{equation}
where $F_{Res} \in \mathbb{R}^{ \frac{W}{32} \times \frac{H}{32} \times 512}$ represents the output of ResNet50. 

\subsubsection{Transformer Learning branch}

Due to partial occlusion of pedestrians, key parts of the body are often missing, making it difficult to accurately identify them. However, the Transformer \cite{bib38, bib39} captures global features, linking these scattered features together and providing a comprehensive pedestrian representation, thereby alleviating the challenges posed by occlusion. Therefore, we design a Transformer-based feature extractor that better understands the overall structure of pedestrian images, maintaining high recognition efficiency.

Firstly,the input pedestrian image I is divided into $N$ fixed-size patches of $p \times p$. Then, each image patch undergoes a linear transformation using a learnable projection matrix $E \in \mathbb{R}^{D}$ to convert it into a fixed-length feature vector \begin{equation}
f_{\text {patch }} \in \mathbb{R}^{D},
\end{equation}
where $D$ is the dimension of the feature vector. All feature vectors of the patches are sequentially arranged, and a learnable classification token (denoted as $f_{cls}$) is appended to integrate the features of all patches, generating a global feature representation.
Learnable positional encoding information $E_{posi}$ is added to each patch feature vector to retain the positional information within the original image. Thus, we get the input feature sequence $X_f \in \mathbb{R}^{(N+1) \times D} $ for the Transformer:
\begin{equation}
X_f=\left[f_1^{\text {patch }} E, f_2^{\text {patch }} E, \ldots, f_N^{\text {patch }} E, f_{c l s}\right]+E_{\text {posi}},
\end{equation}
where $N$ represents the number of patches, and $f_k^{\text patch}$ represents $k-th$ patch. $E$ represents an embedding matrix used to map input feature vectors.
The input feature sequence undergoes layer normalization to obtain the normalized feature sequence $X^N$:
\begin{equation}
    X^N = LayerNorm(X_f).
\end{equation}
The normalized feature sequence $X^N$ is multiplied by three weight matrices $W^Q$, $W^K$, and $W^V$ respectively to obtain the query vector $Q$, key vector $K$, and value vector $V$:
\begin{equation}
\begin{aligned}
& Q=X^N W^Q, \\
& K=X^N W^K, \\
& V=X^N W^V.
\end{aligned}
\end{equation}
The multi-head attention mechanism calculates the relationships between the feature vectors. The output of the multi-head attention is denoted as $Z_H$:
\begin{equation}
Z_H=\operatorname{MultiHead}(Q, K, V)=\operatorname{Concat}\left(H_1, H_2, \ldots, H_h\right) W^O,
\end{equation}
where each head $H_h$ represents $h-th$ single self-attention module computing by the following equation, where $\sqrt{d_k}$ is used to scale $QK^T$ to prevent them from becoming too large, and $W^O$ is the weight matrix for the multi-head outputs.
\begin{equation}
H_h=\operatorname{softmax}\left(\frac{Q_h K_h^T}{\sqrt{d_k}}\right) V_h,
\end{equation}
Utilize T-cascaded encoder models to extract the relationships among the previously obtained input local feature sequences,the final output sequence $F_{Trans} \in \mathbb{R}^{(N+1) \times D} $ includes the classification token $f_{cls}^{out}$, which represents the global feature of the entire image:
\begin{equation}
F_{Trans}=\left[f_1^{\text {out }}, f_2^{\text {out }}, \ldots, f_N^{\text {out }}, f_{\text {cls }}^{\text {out }}\right].
\end{equation}
where $f_1^{\text{out}}$ is the output representation of $f_1^{\text{patch}}$ through the Transformer model, and similarly for the others. Using the multi-head self-attention mechanism of the Transformer encoder, the features from different patch blocks are weighted and fused to construct a global image feature representation. The fused features can then be further applied to GCM to extract more discriminative features.

\subsection{Graph Convolutional Module (GCM)}

To effectively integrate features from different branches, i.e., the affinity field and confidence maps from the pose estimation, the feature maps from ResNet and Transformer, we propose a GCM to combine these features and produce a comprehensive feature representation for Person Re-Identification.

First, for the outputs of pose estimation, the affinity field $L$ and confidence maps $S$ are used to construct node and edge features for the graph. The specific steps are as follows: 1) Node Features Initialization: Initialize node features $H_S \in \mathbb{R}^{k \times j}$ using the confidence maps $S$. Each node corresponds to a key point, and the node features $H_S$ represent the confidence score of that key point as follows:
\begin{equation}
H_S=\operatorname{Normalize}\left(S_k^j\right),
\end{equation}
where $S_{k}^{j}$ denotes the confidence score of the $k$-th key point at the $j$-th location. Edge Feature Initialization: Initialize edge features $H_L \in \mathbb{R}^{k \times k}$ using the affinity field $L$. Each edge represents the degree of association between key points.
\begin{equation}
H_L=\operatorname{Normalize}\left(L_{i j}\right),
\end{equation}
where $L_{ij}$ represents the affinity score between the $i$-th and $j$-th key points.

Second, we process the feature maps from ResNet and Transformer to obtain features suitable for the graph neural network. 
ResNet Feature Maps: Convert the feature map $F_{Res} \in \mathbb{R}^{k \times d_{Res}}$ into node features $H_{Res}$ via convolutional operations. 
\begin{equation}
H_{\text {Res }}=\operatorname{Conv}_{\text {Res }}\left(F_{R e s}\right).
\end{equation}
2) Transformer Feature Maps: Convert the feature map $F_{Trans}$ into node features $H_{Trans} \in \mathbb{R}^{k \times d_{Trans}}$ via convolutional operations.
\begin{equation}
H_{\text {Trans }}=\operatorname{Conv}_{\text {Trans }}\left(F_{\text {Trans }}\right).
\end{equation}

Next, initialize the graph’s node features $H_{node} \in \mathbb{R}^{k \times (j + d_{Res} + d_{Trans})}$ and edge features $H_{edge} \in \mathbb{R}^{k \times k}$ as follows:
\begin{equation}
\begin{gathered}
H_{\text {node }}=\operatorname{Concat}\left(H_S, H_{\text {Res }},H_{\text {Trans }}\right), \\\end{gathered}\end{equation}
and
\begin{equation}
\begin{gathered}
H_{\text {edge }}=\operatorname{Normalize}\left(H_L\right).
\end{gathered}
\end{equation}
where $\text{Concat}$ denotes concatenation of node features. Graph convolution layers are used to update node features by considering edge features.
\begin{equation}
H_{\text {node }}^{(l+1)}=\operatorname{GCM}\left(H_{\text {node }}^{(l)}, H_{\text {edge }}\right),
\end{equation}
where $\text{GCM}(\cdot)$ represents the graph convolution operation and $l$ denotes the layer index. In the GCM, node features are updated and aggregated at each layer. The final node features $H_{agg}$ are obtained through an aggregation operation.
\begin{equation}
H_{\text {agg }}=\operatorname{Aggregate}\left(H_{\text {node }}^{(N)}\right),
\end{equation}
where $\text{Aggregate}$ performs feature aggregation and $N$ is the total number of graph convolution layers.

Finally, we fuse the aggregated node features $H_{agg}$ with other features to obtain the final fused feature $F_{final}$:
\begin{equation}
F_{final}=\operatorname{Fusion}\left(H_{\text {agg }},H_{\text {node }}^{(L)}\right),
\end{equation}
The final fused feature $F_{final}$ is fed into a classifier for Person Re-Identification to get $P_{reid}  \in \mathbb{R}^{C}$:
\begin{equation}
P_{\text {reid }}=\text { Classifier }\left(F_{\text {final }}\right),
\end{equation}
where $\text{Classifier()}$ is a classifier typically consisting of fully connected layers and activation functions,$P_{\text{reid}}$ is the result after classification.

In summary, the proposed graph convolutional module successfully integrates various features from different branches. This approach leverages the information from each feature source to extract more discriminative pedestrian characteristics. This not only improves the model's accuracy in Person Re-Identification tasks but also significantly enhances its robustness in various complex scenarios.Next we discuss Loss Function for the proposed method.

\subsection{Loss Function Design}
To enhance performance, the three branches—pose estimation, ResNet features, and Transformer features—are trained separately and then combined. Each branch uses a distinct loss function tailored to its specific task, allowing for optimized learning and more effective feature integration.

The pose estimation branch uses OpenPose to generate the affinity fields $L$ and confidence maps $S$. The pose estimation network is trained with the L2 loss to optimize the prediction of $L$ and $S$. The L2 loss for the pose estimation branch can be expressed as:
\begin{equation}
\mathcal{L}_{\text {pose }}=\frac{1}{N} \sum_{i=1}^N\left\|L_i-L_i^{\mathrm{gt}}\right\|^2+\frac{1}{N} \sum_{i=1}^N\left\|S_i-S_i^{\mathrm{gt}}\right\|^2,
\end{equation}
where $L_i$ and $S_i$ are the predicted affinity fields and confidence maps, respectively, and $L_i^{gt}$ and $S_i^{gt}$ are the ground truth values. $N$ denotes the number of keypoints or patches. L2 loss minimizes the squared difference between the predicted and ground truth values, improving the accuracy of the pose estimation by ensuring precise localization of keypoints and reliable confidence scores.

The ResNet branch extracts deep features from pedestrian images. It employs a contrastive loss  which encourages similar features to be close and dissimilar features to be far apart in the feature space. The loss function is defined as:
\begin{equation}
\mathcal{L}_{c e}=\frac{1}{M} \sum_{i=1}^M\left[y_i \cdot \max \left(0, \operatorname{margin}-\left\|F_i^{\mathrm{res}}-F_i^{\mathrm{pos}}\right\|^2\right)+\left(1-y_i\right) \cdot\left\|F_i^{\mathrm{res}}-F_i^{\mathrm{neg}}\right\|^2\right],
\end{equation}
where $F_i^{\mathrm{res}}$ represents the feature vector from ResNet, $F_i^{\mathrm{pos}}$ and $F_i^{\mathrm{neg
}}$ are positive and negative pairs respectively, $y_i$ is a binary label indicating similarity, and 
margin is a hyperparameter that defines the minimum distance between dissimilar pairs. Contrastive loss enhances the discriminative power of the ResNet features by ensuring that features from the same identity are close while features from different identities are separated, improving recognition performance.

The Transformer branch captures global contextual information using self-attention mechanisms. It uses triplet loss to enforce a relative distance constraint between anchor, positive, and negative samples:
\begin{equation}
\mathcal{L}_{\text {triplet }}=\frac{1}{P} \sum_{i=1}^P \max \left(0,\left\|F_i^{\mathrm{trans}}-F_i^{\mathrm{pos}}\right\|^2-\left\|F_i^{\mathrm{trans}}-F_i^{\mathrm{neg}}\right\|^2+\text { margin }\right),
\end{equation}
where $F_i^{\mathrm{trans}}$ represents the feature vector from ResNet. Triplet loss ensures that the global features learned by the Transformer are sufficiently discriminative by maintaining the desired relative distances between feature embeddings, which is crucial for accurate Person Re-Identification.

Once trained, the features from these branches are concatenated or fused to create a unified representation. The combined feature vector incorporates:
\begin{equation}
    F_{Combined} = F_{pose} + F_{trans} + F_{res}.
\end{equation}
where $F_{pose}$ is derived from pose estimation features, $F_{res}$ is from ResNet features, and $F_{trans}$ is from Transformer features.

Training each branch with a specific loss function allows the model to leverage different types of information—local features, contextual understanding, and global representations. This comprehensive learning approach enhances the model's ability to extract discriminative features and improve performance in Person Re-Identification tasks. By separately optimizing each branch for its specific role and then integrating their outputs, the model achieves higher accuracy and robustness in various scenarios.

\section{Experiments and Analysis}
To verify the superiority of the proposed Tran-GCN method, experiments are conducted from both quantitative and qualitative angles. Specifically, ablation experiments and comparative experiments are adopted for quantitative analysis. In the comparative experiments, the Market-1501, DukeMTMC-ReID, and MSMT17 datasets commonly used in Person Re-Identification are employed to evaluate the proposed model.
\subsection{Dataset}
The Market-1501 dataset comprises 32,668 pedestrian images of 1,501 unique identities, captured by six cameras with different viewpoints on the Tsinghua University campus. The pedestrian bounding boxes in all images are extracted from the original video frames using the DPM \cite{bib40} pedestrian detector. The training set contains 12,936 images from 751 identities, with an average of 71.2 images per identity. The test set includes 15,913 gallery images and 3,368 query images from 750 identities.

The DukeMTMC-ReID dataset provideds a large dataset recorded by eight cameras, which included 36,411 labeled images of 1404 identities. The 1404 identities are randomly divided, with 702 identities for training and the others for testing. Among them, 16,522 images of 702 identities are used for training, and 2228 query images and 17,661 gallery images of 1110 identities for testing \cite{bib41}.

The MSMT17 dataset is the most recent and largest Person Re-Identification dataset that closely resembles real-world scenarios. It contains 126,441 images of 4,101 identities, captured by 15 cameras in a campus environment. The pedestrian bounding boxes are detected from the original video frames using the Faster-RCNN detector \cite{bib42}. The training set comprises 32,621 images from 1,041 identities, with an average of approximately 31.3 images per identity. The test set consists of 82,161 gallery images and 11,659 query images.

\subsection{ Experimental Environment}
The primary environment is as follows: Python 3.7, PyTorch 1.6,64-bit Ubuntu 18.04 operating system, CPU: Xeon E5-2620, Memory: 32GB, GPU: NVIDIA GTX 2080Ti with 12GB VRAM, CUDA version: 11.1.

\subsection{Evaluation Metrics}

This paper uses two person re-identification evaluation metrics to evaluate the proposed model:

(1) Rank-k Hit Rate (Rank-k). It represents the probability that the target to be retrieved appears within the top k positions of the retrieval results. In this paper, Rank-1, Rank-5, and Rank-10 are selected as evaluation metrics to measure the model's performance, with Rank-1 serving as an important reference. A higher Rank-1 value indicates a higher hit rate at the first position, signifying better model performance.

(2) Mean Average Precision (mAP). To provide a more comprehensive assessment of the model's overall performance, the mean Average Precision (mAP) is utilized as an evaluation criterion. mAP considers the positions of all images with the same ID as the retrieval target within the retrieval results. A higher mAP value indicates that the correct retrieval targets are ranked higher in the results, reflecting the algorithm's average accuracy performance effectiveness of the Tran-GCN model.

Cumulative Matching Characteristic (CMC) curve and mean Average Precision (mAP). CMC demonstrates the accuracy of the top K individuals by calculating the true positives and false positives among the top K individuals in the ranked list. mAP measures the area under the precision-recall curve, reflecting the overall re-identification accuracy across the gallery set \cite{bib43}.

For the Tran-GCN performance is evaluated quantitatively by mean average precision (mAP) and cumulative matching characteristic (CMC) at Rank-1, Rank-5, Rank-10 \cite{bib44}.

\subsection{Ablation Experiments}

To validate the effectiveness of the proposed Tran-GCN, we conducted ablation experiments on the Market-1501 dataset as shown in Table \ref{tab:tb1}. We employed different methods, including Baseline, GCM (without the addition of Transformer), and Tran-GCN, to derive the respective values for the evaluation metrics of Rank-1, Rank-5, Rank-10, and mAP.From Table \ref{tab:tb1}, the ResNet-50 baseline method achieved an accuracy of 92.1\% for Rank-1 and 81.7\% for mAP. In contrast, the GCM improved performance significantly, achieving 96.4\% for Rank-1 and 87.1\% for mAP.When the Transformer was added to the GCM Module(Tran-GCN), the accuracy for Rank-1 and mAP improved to 97.4\% and 87.9\% respectively, representing an increase of 1.0\% for mAP and 0.8\% for Rank-1 over the GCM model, and an increase of 5.3\% and 6.2\% respectively over the baseline for Rank-1 and mAP.Thus,Tran-GCN can better extract local features and the global dependencies between feature sequences, and can further improve the recognition accuracy of the model.

\begin{table}[!t]
  \caption{Results of ablation experiments on the Market-1501 dataset.}
  \label{tab:tb1}
  \centering
  \begin{tabular}{@{}lllll@{}}
    \toprule
    Method & Rank-1 & Rank-5 & Rank-10 & mAP \\
    \midrule
    Baseline & 92.0 & 95.9 & 98.0 & 81.6 \\
   \textbf{GCM} & \textbf{96.2} & \textbf{98.0} & \textbf{98.8} & \textbf{86.9} \\
    \textbf{Tran-GCN} & \textbf{97.2} & \textbf{98.4} & \textbf{99.0} & \textbf{87.7}\\
  \bottomrule
  \end{tabular}
\end{table}

\begin{figure}[!t]
  \centering
  \includegraphics[height=6.2cm,width=11.2cm]{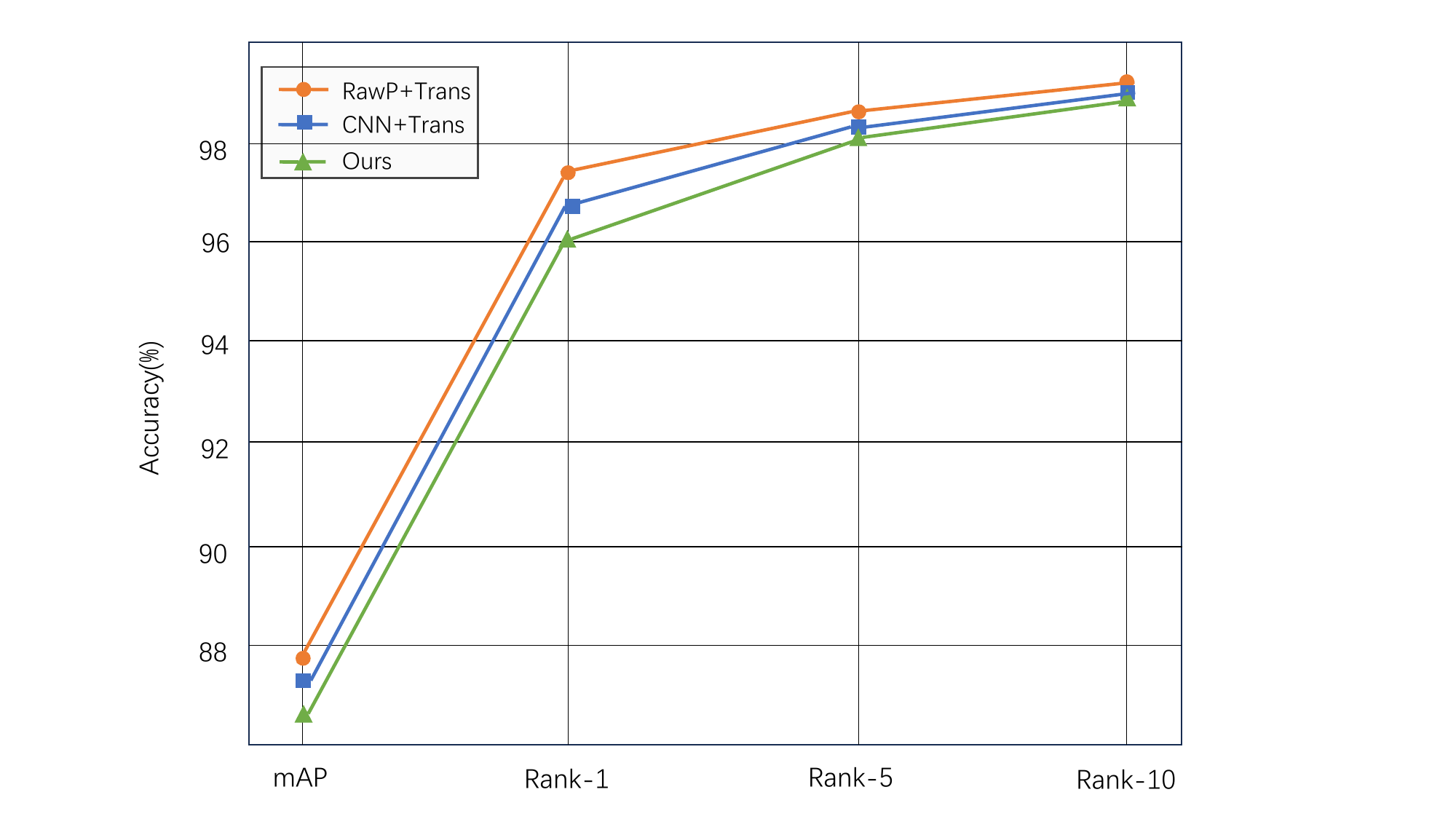}
  \caption{Experimental results of different partitioning methods for Transformer input.)}
  \label{fig:fig2}
\end{figure}

To verify the effectiveness of the constructed fine-grained and semantically localized features within the Transformer learning branch, we conducted ablation experiments on the Market-1501 dataset. This experiment employed three methods: RawP+Trans, CNN+Trans, and Ours. The three methods were executed under identical Transformer network parameters and experimental conditions, and the experimental results are presented in Fig. \ref{fig:fig2}. 
Among the methods, the RawP+Trans approach directly divides the original pedestrian image horizontally and vertically to obtain feature sequences for each image patch, which are then input into the Transformer encoder model. The CNN+Trans method uses ResNet-50 to extract the pedestrian feature map, where the vector at each pixel location in the feature map is taken as the input feature sequence. After flattening and linear transformation, the input feature sequence is fed into the Transformer. Our method utilizes the first two stages of ResNet-50 to extract fine-grained pedestrian features, leveraging keypoint information to extract semantically localized features. These features are then flattened, concatenated, and linearly transformed to obtain the input features for the Transformer. As shown in Fig. \ref{fig:fig2}, our proposed method achieves an accuracy of 87.9\% for mAP and 97.4\% for Rank-1. This approach not only retains detailed features from the original image but also extracts more semantically localized regional features of pedestrians, resulting in more discriminative pedestrian features and higher recognition accuracy.

\subsection{Comparative Experiments}

To further verify the effectiveness of the Tran-GCN model proposed in this paper, we conducted comparative experiments on three public datasets (Market-1501, DukeMTMC-ReID, and MSMT17).

From table \ref{tab:tb2}, it shows the comparative experimental results of the Tran-GCN model on the Market1501 dataset. The compared methods include part-based methods (PCB, PCB+RPP, AlignedReID, MGN, Deep-pedestrian), mask-based methods (MGCAM, MaskReID, SPReID), and pose-based methods (SpindleNet, PIE, PDC, PAR, PSE, etc.). Tran-GCN achieves accuracies of 97.4\%, 98.4\%, 99.0\%, and 87.9\% for Rank-1, Rank-5, Rank-10, and mAP, respectively, on this dataset, surpassing most mainstream methods. This indicates that Tran-GCN can effectively aggregate the global dependencies among fine-grained local features of pedestrians, enabling the model to focus on more important local regions and further improving recognition accuracy.

\begin{table}[!t]
  \caption{Comparative Experimental Results of Tran-GCN on Market-1501 Dataset.}
  \label{tab:tb2}
  \centering
  \begin{tabular}{p{3.3cm}cccccc}
    \toprule
        & \textit{Method} & \textit{Rank-1} & \textit{Rank-5} & \textit{Rank-10} & \textit{mAP} \\
    \midrule
    \multirow{5}{*}{Part-based methods} & PCB\cite{bib8} & 92.3 & 97.2 & 98.2 & 77.4 \\ 
    & PCB+RPP\cite{bib8} & 93.8 & 97.5  & 98.5 & 81.6 \\
    & AlignedReID\cite{bib9} & 91.8 & 97.1 & - 79.2 \\
    & MGN\cite{bib5} & 95.7 & - & - & 86.9 \\
    & Deep-pedestrian\cite{bib26} & 92.3 & - & - & 79.6 \\
    \midrule
    \multirow{3}{*}{Mask-based methods} & MGCAM\cite{bib6} & 83.8 & - & - & 74.3 \\ 
    & MaskeReID\cite{bib29} & 90.0 & - & - & 75.3 \\
    & SPReID\cite{bib28} & 92.5 & 97.2 & 98.1 & 81.3 \\
    \midrule
    \multirow{10}{*}{Pose-based methods} & SpindleNet\cite{bib7} & 76.9 & 91.5 &  94.6 & - \\ 
    & PIE\cite{bib13} & 78.7 & 90.3 & 93.4 & 53.9 \\
    & PDC\cite{bib14} & 84.1 & 92.7 & 94.9 & 63.4 \\
    & PAR\cite{bib32} & 81.0 & 92.0 & 94.7 & 63.4 \\
    & PSE\cite{bib34} & 87.1 & - & - & 69.0 \\
    & Part-Aligned\cite{bib15} & 91.7 & - & - & 79.6 \\
    & PGR\cite{bib45} & 93.8 & 97.7 & - & 77.2 \\
    & PGFA\cite{bib46} & 91.2 & - & - & 76.8 \\
    & Pose-transfer\cite{bib47} & 87.7 & - & - & 68.9\\
    & PN-GAN\cite{bib48} & 89.4 & - & - & 72.6\\
    \midrule
    \multirow{2}{*}{} & \textbf{GCM} & 96.3 & 98.1 & 98.9 & 87.0 \\ 
    & \textbf{Tran-GCN} & \textbf{97.2} & \textbf{98.4} & \textbf{99.0} &  \textbf{87.7}\\
  \bottomrule
  \end{tabular}
\end{table}

From Table \ref{tab:tb3}, it presents the comparative experimental results of the Tran-GCN model on the DukeMTMC-ReID dataset. It indicates that Tran-GCN outperforms most mainstream methods on this dataset, achieving accuracies of 88.3\%, 93.8\%, 96.0\%, and 78.2\% for Rank-1, Rank-5, Rank-10, and mAP, respectively. Its performance is comparable to  the MGN method, which is known for its high accuracy among horizontal partition-based methods. This proves the effectiveness of incorporating the Transformer Learning Module.

\begin{table}[!t]
  \caption{Comparative Experimental Results of Tran-GCN on  DukeMTMC-ReID Dataset.}
  \label{tab:tb3}
  \centering
  \begin{tabular}{p{3.3cm}cccccc}
    \toprule
        & \textit{Method} & \textit{Rank-1} & \textit{Rank-5} & \textit{Rank-10} & \textit{mAP} \\
    \midrule
    \multirow{5}{*}{Part-based methods} & PCB\cite{bib8} & 81.7  & -  & - & 66.1  \\
    & PCB+RPP\cite{bib8}    & 83.3   & -  & - & 69.2  \\
    
    & MGN\cite{bib5}    & 88.7  & -  & - & 78.4 \\
    &Deep-pedestrian\cite{bib26} & 80.9 & - & - & 64.8\\
    \midrule
    \multirow{3}{*}{Mask-based methods}    &  MaskReID\cite{bib29}    & 78.7   & -  & - & 61.9 \\
    &SPReID\cite{bib28} & 85.9 & 92.9 & 94.5 & 73.3\\
    \midrule
    \multirow{10}{*}{Pose-based methods} &  PSE\cite{bib34}  & 79.8   & 89.7  & 92.2 & 62.0  \\
&Part-Aligned\cite{bib15}    & 84.4   & -  & - & 69.3 \\
&PGR\cite{bib45}    & 83.6  & 89.7  & 92.2 & 65.9 \\
&PGFA\cite{bib46}    & 82.6  & -  & - & 76.8 \\
&Pose-transfer\cite{bib47}    & 78.5   & -  & - & 56.5  \\
&PN-GAN\cite{bib48}    & 73.6   & -  & - & 53.2 \\
    \midrule
    \multirow{2}{*} & \textbf{GCM}    & \textbf{87.5}   &\textbf{93.3}  & \textbf{95.6} & \textbf{77.1}  \\
 &\textbf{Tran-GCN}    & \textbf{88.3}   & \textbf{93.8}  & \textbf{96.2} & \textbf{78.2} \\
  \bottomrule
  \end{tabular}
\end{table}

Table \ref{tab:tb4} displays the comparative experimental results of Tran-GCN on the even larger MSMT17 dataset. The proposed method Tran-GCN improves the accuracies of Rank-1 and mAP by 1.8\% and 2.3\% than GCM, respectively, and surpasses most classical algorithms. This proves the effectiveness and advanced nature of Tran-GCN on large-scale datasets.

\begin{table}[!t]
\caption{Comparative Experimental Results of Tran-GCN on MSMT17 Dataset(\%).}\label{tab:tb4}%
\begin{tabular}{@{}lllll@{}}
\toprule
\textit{Method}  & Rank-1   & Rank-5 & Rank-10  & mAP \\
\midrule
GoogLeNet\cite{bib49}    & 47.6   & -  & - & 23.0  \\
PDC\cite{bib14}    & 58.0   & -  & - & 29.7  \\
GLAD\cite{bib33}    & 61.4   & -  & - & 34.0 \\
PCB+RPP\cite{bib8} & 68.2 & - & - & 40.4\\
MGN\cite{bib5} & 76.9 & - & - & 52.1\\
\midrule
\textbf{GCM}    & \textbf{78.4}   & \textbf{88.5}  & \textbf{91.3} & \textbf{54.3}  \\
\textbf{Tran-GCN}    & \textbf{80.2}   & \textbf{89.6}  & \textbf{92.2} & \textbf{56.6} \\
\botrule
\end{tabular}
\end{table}

\begin{figure}[!t]
  \centering
  \includegraphics[width=1\linewidth]{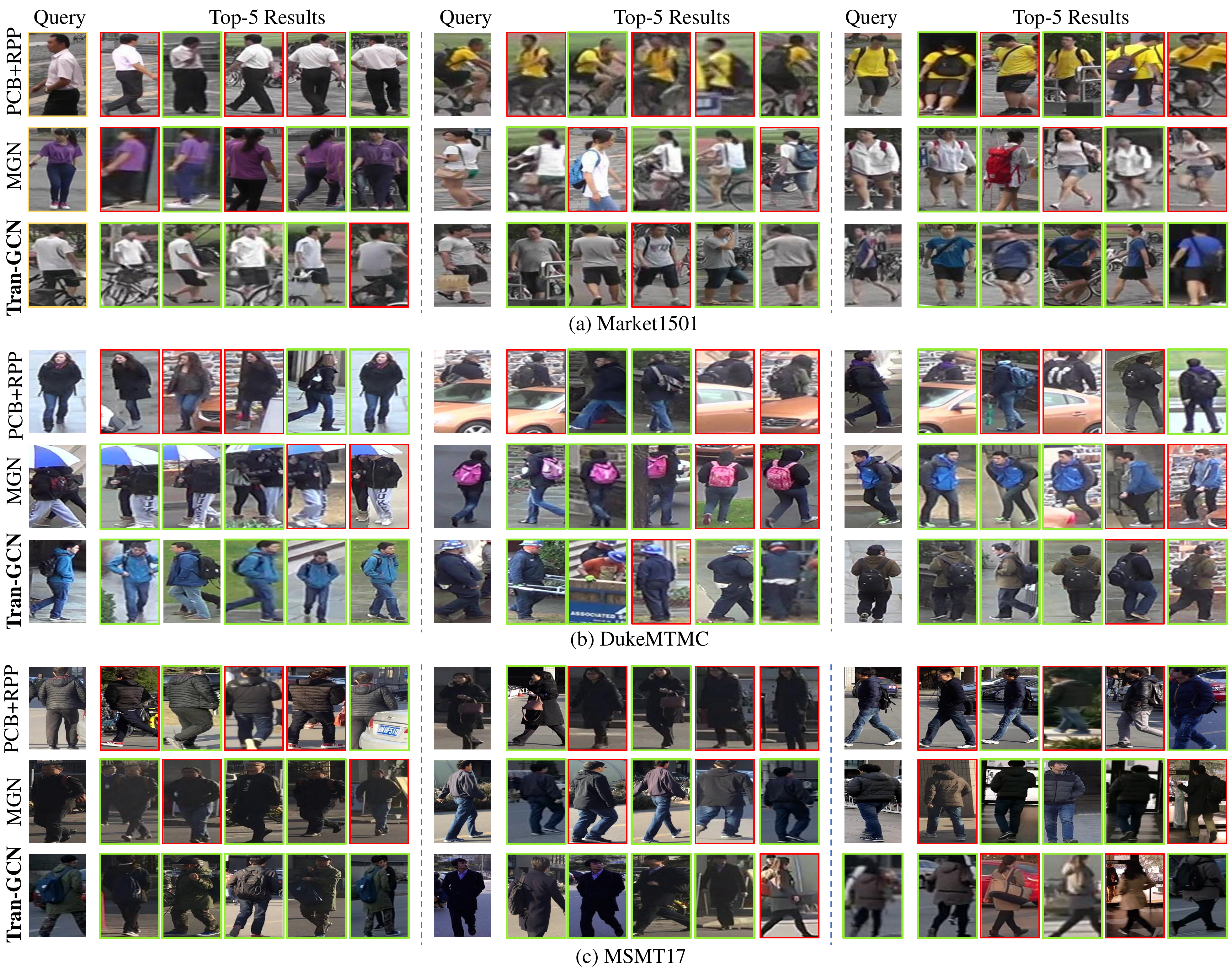}
  \caption{Visualization of top-5 retrieval results of our Tran-GCN method on three datasets.}
  \label{fig:fig3}
\end{figure}

From Fig. \ref{fig:fig3}, the experimental results of the Tran-GCN method on three public retrieval datasets can be clearly observed. On the left side of the figure, the query pedestrian images are displayed, while the right side shows the Top-5 retrieval results, with correct results marked in green and incorrect results marked in red. It can be seen that even in challenging scenarios with complex backgrounds, similar pedestrian appearances, and partial occlusions in the images, Tran-GCN still achieves satisfactory retrieval performance. Particularly in these complex situations, Tran-GCN demonstrates robustness and efficiency in handling visual retrieval tasks. These results indicate that the Tran-GCN method has potential in practical applications, effectively enhancing the accuracy and reliability of visual retrieval.

\section{Discussion}
Despite the Tran-GCN model's ability to integrate multiple feature types and improve pedestrian re-identification accuracy, it is not without its limitations.

\noindent 
\textbf{Computational Complexity and Resource Demands}. The Tran-GCN model combines the outputs from OpenPose, ResNet-50, and Transformer encoders. Each of these components is computationally intensive, requiring significant processing power and memory. This complexity can pose challenges for real-time applications and deployment on devices with limited resources.

\noindent
\textbf{Training Time and Data Requirements.} The multi-branch learning approach, involving separate training of each branch before integration, can lead to prolonged training times. Additionally, the model's reliance on extensive labeled datasets, such as those used for OpenPose, ResNet, and Transformer training, may limit its applicability in scenarios where such comprehensive datasets are unavailable.

\noindent
\textbf{Model Interpretability.}
As the model integrates features from various sources and applies complex transformations, interpreting its decision-making process becomes challenging. Understanding how different features contribute to the final re-identification output is crucial for improving model transparency and trustworthiness.

\section{Conclusion}

In this study, we propose a Person Re-Identification model based on the Transformer-Enhanced Graph Convolutional Neural Network (Tran-GCN) to improve pedestrian recognition in monitoring videos. The Tran-GCN model consists of a multi-branch feature extraction module and a graph convolutional network module(GCM). The multi-branch feature extraction includes pedestrian keypoint features, local features, and global features. The GCM integrates these three different types of features to obtain a discriminative feature representation. Extensive experiments have demonstrated that the proposed method effectively enhances the accuracy of Person Re-Identification. In the future, we plan to further integrate mutual information for multi-feature fusion and extend the application to challenging scenarios involving low-light conditions and disguised pedestrians.

\section{Acknowledgments}\label{sec13}

The first author would like to thank you for the support of the College of Guangzhou technology and business of China. We also want to thank the HCI V3Lab of Universiti Teknologi Malaysia for some constructive comments. 

\section*{Declarations}

\textbf{Authors’ contributions\textit{\newline
}}Masitah put up with the idea,Tarmizi guided the research process and checked the work, Hong xiaobin designed the model,done the experiment and wrote the main manuscript. All authors reviewed the manuscript.
\noindent

\bigskip

\begin{appendices}




\end{appendices}


\bibliography{sn-article}

\end{document}